\documentclass[preprint,12pt]{elsarticle}

\usepackage{amssymb}
\usepackage{amsmath}
\usepackage[ruled,vlined]{algorithm2e}

\journal{Computer Vision and Image Understanding}

\begin{document}

\begin{frontmatter}



\title{OVGrasp: Open-Vocabulary Grasping Assistance via Multimodal Intent Detection}


\author[label1]{Chen Hu} 
\author[label2]{Shan Luo}
\author[label1,label3]{Letizia Gionfrida\corref{cor1}}
\cortext[cor1]{Corresponding author}
\affiliation[label1]{organization={Department of Informatics, King's College London},
            addressline={30 Aldwych}, 
            city={London},
            postcode={WC2R 4BG}, 
            state={},
            country={uk}}
\affiliation[label2]{organization={Department of Engineering, King's College London},
            addressline={30 Aldwych}, 
            city={London},
            postcode={WC2R 4BG}, 
            state={},
            country={uk}}
\affiliation[label3]{organization={John A. Paulson School of Engineering and Applied Sciences, Harvard University},  
            addressline={29 Oxford Street}, 
            city={Cambridge},
            postcode={02138},
            state={MA 02138},
            country={USA}}

\begin{abstract}
Grasping assistance is essential for restoring autonomy in individuals with motor impairments, particularly in unstructured environments where object categories and user intentions are diverse and unpredictable. We present \textbf{OVGrasp}, a hierarchical control framework for soft exoskeleton-based grasp assistance that integrates RGB-D vision, open-vocabulary prompts, and voice commands to enable robust multimodal interaction. To enhance generalization in open environments, OVGrasp incorporates a vision-language foundation model with an open-vocabulary mechanism, allowing zero-shot detection of previously unseen objects without retraining. A multimodal decision-maker further fuses spatial and linguistic cues to infer user intent, such as grasp or release, in multi-object scenarios. We deploy the complete framework on a custom egocentric-view wearable exoskeleton and conduct systematic evaluations on 15 objects across three grasp types. Experimental results with ten participants demonstrate that OVGrasp achieves a grasping ability score (GAS) of $87.00 \pm 2.46\%$, outperforming state-of-the-art baselines and achieving improved kinematic alignment with natural hand motion. 

\end{abstract}

\begin{graphicalabstract}
\end{graphicalabstract}

\begin{highlights}
\item Research highlight 1
\item Research highlight 2
\end{highlights}


\end{frontmatter}



\section{Introduction}
Every year, in the United Kingdom alone, approximately 100,000 people experience a stroke annually, equivalent to one stroke every five minutes \cite{StrokeAssociation2023}. Currently, around 1.3 million stroke survivors are living in the UK, making stroke one of the leading causes of disability and the fourth most common cause of death \cite{PublicHealthEngland2021}. Effective hand function, especially the ability to grasp and manipulate objects, is critical for performing activities of daily living (ADLs) and maintaining independent living \cite{takahashi2008, bae2015, hsieh2018comparison}. However, stroke frequently results in severe motor impairments, particularly affecting upper limb functionality and significantly diminishing an individual’s ability to execute grasping and object manipulation tasks \cite{kim2016effects, bland2008restricted}. Despite intensive rehabilitation, many stroke survivors experience persistent upper limb dysfunction, severely restricting autonomy and participation in daily activities. Recent research has focused on assisting grasping and other hand functions using soft wearable robots \cite{gionfrida2024wearable}. Soft hand exoskeletons, typically fabricated using flexible materials such as textiles and silicone, employ compliant mechanisms including pneumatic actuation and tendon-driven systems to facilitate grasping tasks \cite{tanczak2024soft}.

A critical challenge in controlling soft hand exoskeletons is intent detection \cite{gionfrida2024wearable, kim2019eyes}. Accurately interpreting user intent to grasp or release objects is essential to providing intuitive and adaptive assistance. Vision-based intent detection, an emerging context-aware strategy for wearable robotic systems, has received increasing attention as it mirrors the natural human action-triggering process—visually perceiving the environment, processing information through the neural system, and initiating actions. Current context-aware approaches for soft exoskeleton control mainly include geometry-based heuristic methods \cite{tricomi2023environment, diaz2016strong, perez2015detection} and data-driven methods \cite{kim2019eyes, rho2024multiple, rho2021learning}. Geometry-based heuristic methods do not require training on specific datasets and directly analyze the environment within the camera’s field of view, such as objects placed on a table or stairs for climbing tasks. Although computationally efficient and highly generalizable, these methods possess limited capability for interpreting complex, open-ended environments and typically rely on computationally expensive reconstruction of 3D object information, such as point clouds \cite{tricomi2023environment}. data-driven methods \cite{rho2021learning}, conversely, train deep neural networks on extensive datasets or fine-tune pretrained models on specific datasets. Despite their demonstrated effectiveness, these methods remain constrained to detecting objects within predefined categories (e.g., the 80 classes in the COCO dataset \cite{lin2014microsoft}). Once trained, the detectors can only recognize these fixed categories, severely limiting their generalizability and applicability in open-ended scenarios \cite{cheng2024yoloworld}.

\begin{figure*}[t]
\centering
\includegraphics[width=0.9\textwidth]{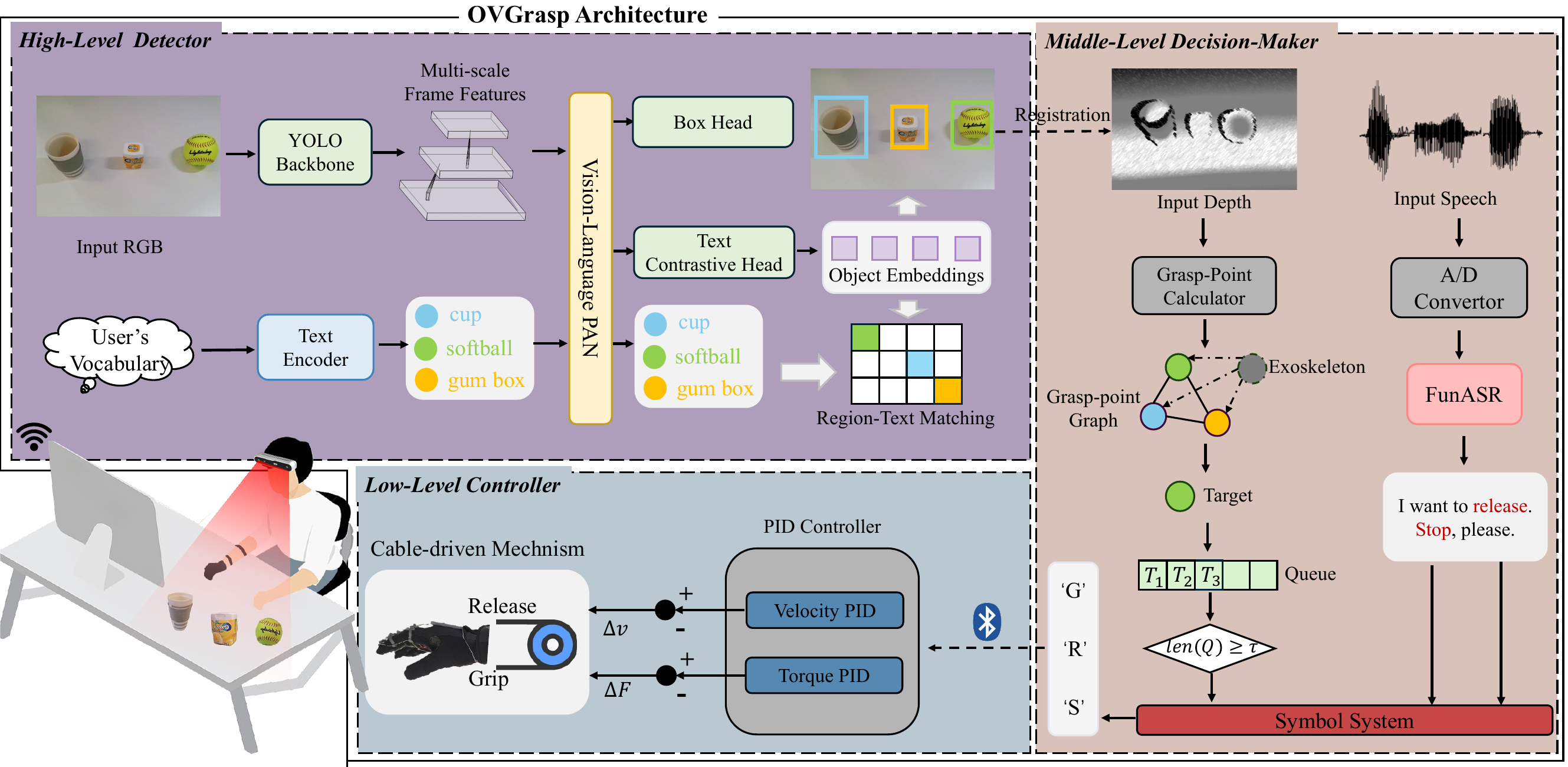}
\caption{\textbf{An Overview of OVGrasp Architecture.} This hierarchical control framework consists of three primary components. \textbf{1) High-Level Detector:} This module, based on an existing design \cite{cheng2024yoloworld}, employs a YOLO-based visual backbone and a text encoder to generate multi-scale visual and textual embeddings. These embeddings are then integrated using a Vision-Language Path Aggregation Network (PAN). A Box Head and a Text Contrastive Head further process these embeddings to achieve open-vocabulary object detection through region-text matching. \textbf{2)Middle-Level Decision-Maker:} This module integrates depth information and speech commands to perform multimodal intent detection. A grasp-point position graph of detected objects is constructed through depth-image registration. Distances between the hand exoskeleton and each object are calculated in real-time, identifying the nearest object as the target. Concurrently, interactive voice commands processed by the FunASR \cite{gao2023funasr} speech recognition system are integrated. Both modalities feed into a symbolic system to confirm grasp or release intentions. \textbf{3)Low-Level Controller:} This component translates decisions into precise motor commands using a PID-based feedback control system. It regulates the cable-driven mechanism of the exoskeleton glove through three-loop control (angular position, velocity, and torque).
}
\label{fig:overview}
\end{figure*}

Recent studies \cite{du2022learning, shi2023edadet, wu2023aligning} have explored vision-language foundation models (VLMs), such as YOLO-World \cite{cheng2024yoloworld}, leveraging language encoders to achieve open-vocabulary object detection. By embedding language modules into vision models, VLMs more closely emulate human cognitive processes, thus providing superior generalization and interpretation capabilities in complex environments. The introduction of open-vocabulary mechanisms significantly enhances zero-shot recognition capability in open scenarios. Specifically, users can introduce new object categories by simply adding corresponding textual descriptions into the prompt list without the need for traditional fine-tuning on new data. However, while VLMs offer powerful generalization, their unrestricted detection capability often results in numerous irrelevant detections (e.g., tables, cabinets, ovens) or impractical items (e.g., drills, hammers) that stroke survivors rarely use, significantly complicating intent detection. Thus, current grasp-assistive systems based on VLMs require customization capabilities tailored to specific user tasks and environments.

Vision-based grasp assistance currently faces two additional significant challenges: 1) The single-intent triggering issue, where current visual-based methods typically trigger only a grasp intent autonomously but fail to effectively detect and trigger release intent once the user grasps and holds an object. \cite{hu2025pointcloudbasedgraspingsoft, hu2025multiclearmultimodalsoftexoskeleton} 2) Intent decision-making in multi-object scenarios remains difficult, as existing methods predominantly classify single objects independently, without effectively handling multiple objects within realistic environments.

Existing vision-based assistive grasping systems typically employ either head-mounted (egocentric) or hand-mounted (eye-in-hand) camera perspectives \cite{missiroli2023integrating, kim2019eyes}. Eye-in-hand \cite{hu2025pointcloudbasedgraspingsoft} setups provide structural simplicity and convenient integration into wearable devices, but limited camera perception at close distances can adversely impact intent detection accuracy, particularly in cluttered scenarios. In addition, the downward-facing hand-mounted camera may physically collide with the table surface during grasp attempts, disturbing the natural hand movements of users \cite{hu2025pointcloudbasedgraspingsoft}. Egocentric systems \cite{kim2019eyes} effectively address this limitation but increase overall complexity and inconvenience for daily use. Importantly, a systematic quantitative comparison between egocentric and eye-in-hand views in assistive grasping has yet to be reported in the existing literature.

To address these challenges, this study proposes OVGrasp, with the following contributions:

\begin{enumerate}
\item We propose \textbf{OVGrasp}, a hierarchical control framework for grasp assistance using wearable hand exoskeletons. OVGrasp leverages multimodal inputs and integrates a vision-language foundation model for enhanced scene understanding and user interaction.

\item To enhance the generalizability in open and dynamic environments, we incorporate an open-vocabulary mechanism within OVGrasp, enabling contextual understanding of unseen objects and scenarios.

\item We introduce a multimodal decision-making module as an intermediate control layer within OVGrasp. This module fuses visual, depth, and speech modalities, detecting users' multiple intentions in real-time for the improvement of interactive control capability in complex environments.

\item OVGrasp is deployed into a custom-designed cable-driven soft hand exoskeleton with an egocentric RGB-D sensor. Extensive experiments, including evaluations using the grasp ability score and quantitative analysis of finger joint kinematics, demonstrate that OVGrasp outperforms state-of-the-art methods, confirming its effectiveness for real-world assistive grasping scenarios.

\end{enumerate}

\section{Related work}
\subsection{Hand Exoskeleton}
Hand exoskeleton devices are classified into rigid and soft categories \cite{du2021review}. Traditional rigid exoskeleton devices \cite{10587178} are often fabricated from stiff materials, providing robustness and durability essential for precise and highly repeatable movements in neuromuscular training involving substantial forces. However, this rigidity often renders such devices bulky, restricting their usability in clinical settings and making them unsuitable for mild impairments or prolonged daily wear \cite{tanczak2024soft}. Drawing inspiration from nature, integrating soft, elastic components into rigid mechanical structures has been proposed to facilitate naturalistic movements consistent with human motor control principles. In contrast, soft exoskeletons \cite{alicea2021soft} are lightweight, flexible, and naturally conform to human movements, providing a more personalized and comfortable fit \cite{tanczak2024soft}. By being less intrusive and more adaptable, these devices offer sustained, gentle assistance for individuals with mild to moderate impairments without environmental constraints.

Soft hand exoskeletons are typically categorized according to their actuation methods into pneumatic and cable-driven systems \cite{tanczak2024soft}. Pneumatic systems \cite{ge2020design} utilize compressed air to inflate elastomeric chambers that, depending on their geometry, can bend, extend, contract, or even elongate. Although this approach remains highly popular, the required compressed-air supply often affects portability. As an alternative approach to achieving compliance, cable-driven systems \cite{hu2025pointcloudbasedgraspingsoft} transmit force by applying tension through strategically placed anchors embedded within textiles to facilitate motion. Due to the compact size of winding motors, these designs can be highly compact; however, cable-driven transmission poses challenges such as friction, cable slack, and backlash. Additionally, spring-blade mechanisms have been proposed, which translate linear motion inputs into rotational motion to generate the forces necessary for bending and extending fingers in hand exoskeletons \cite{tanczak2024soft}. To further address these challenges, we designed a set of customized 3D-printed components integrated onto an exoskeleton glove. This approach maximizes glove softness and comfort, while the lightweight, wear-resistant PLA components significantly reduce friction associated with cable-driven transmissions.

\subsection{Intent Detection Controller}
Recent advancements in upper-limb wearable robots have introduced various control strategies to enhance user interaction, adaptability, and functionality \cite{triwiyanto2021review, du2021review}. Traditional control approaches, such as joystick-based systems, have been widely used \cite{alicea2021soft, sierotowicz2022emg}. However, these systems often suffer from limitations in dexterity and are not well-suited for executing complex or nuanced movements. However, push-button control systems often suffer from limited dexterity, restricting the user's ability to execute adaptive, or multi-degree-of-freedom movements \cite{alicea2021soft}. These systems typically offer discrete, pre-defined actuation modes, making them less effective for tasks that require nuanced interactions. Force-feedback controllers have proven effective in telemanipulation applications by delivering haptic feedback \cite{baselli2024tendon}. However, these systems require physical contact with an object before actuation can occur, struggling to adapt across diverse tasks, such as to provide task-specific assistance \cite{baselli2024tendon}. Control sensors utilizing neuromuscular interfaces have garnered significant attention for their ability to decode user intent from muscle activity, enabling proportional and adaptive assistance \cite{sierotowicz2022emg, lotti2020intention}. Despite their promise, challenges persist in optimizing their integration to achieve more seamless interactions with the environment. 

To address environmental variations, recent advances have focused on enabling wearable robots to perceive and adapt to the user’s surrounding context\cite{tricomi2023environment, gionfrida2024wearable}. By incorporating visual data, these systems can dynamically adjust assistance levels and anticipate user actions, improving adaptability and future planning. Compared to other methods, visual perception has been shown to reduce variability across users and trials while enhancing system responsiveness \cite{tricomi2023environment, kim2019eyes}. However, integrating vision-based systems into wearable devices remains challenging due to their computational demands and reliance on large, application-specific datasets to train deep learning models \cite{zhang2022deep}. Data-driven approaches often struggle in scenarios where training data is limited or unrepresentative of real-world conditions. 

\subsection{Open-Vocabulary Object Detection}
Open-vocabulary object detection (OVD) has recently emerged as a promising direction in modern object detection, aiming to detect and recognize objects beyond predefined category sets. Early approaches to OVD trained detectors primarily on base classes and assessed their performance on novel, unseen categories \cite{zareian2021open, gu2022open}. While these initial methods successfully demonstrated the potential to recognize novel objects, their generalization capabilities remained constrained due to limited training vocabulary and data diversity. To extend object categories and generalization capabilities, recent methods have leveraged large-scale vision-language pre-training (VLP) \cite{radford2021learning}. Notable approaches, such as Detic \cite{zhou2022detecting}, integrate image classification datasets to expand detection vocabulary significantly. OWL-ViT \cite{minderer2022simple} simplifies open-vocabulary detection by fine-tuning vision transformers with automatically labeled datasets. Additionally, approaches like GLIP \cite{li2022grounded} and Grounding DINO \cite{liu2023grounding} utilize phrase grounding frameworks combined with transformer architectures, further enhancing zero-shot capabilities. Recent region-text matching approaches \cite{yao2022detclip} have unified detection, grounding, and image-text tasks into a single training framework, utilizing large-scale datasets to achieve impressive generalization performance. Despite these advancements, these frameworks typically employ large models with heavy backbones, such as Swin Transformers \cite{liu2021swin}, which increase computational complexity and hinder practical deployment.

To address these limitations, YOLO-World \cite{cheng2024yoloworld} presented a lightweight, real-time architecture specifically tailored for efficient open-vocabulary detection. It leverages a re-parameterizable vision-language path aggregation network (RepVL-PAN) that effectively integrates text and image embeddings, significantly improving efficiency without compromising performance. Building upon these works, this study proposes a control paradigm for wearable upper-limb robots that leverages a Vision-Language Foundation Model to interpret the surrounding environment and provide adaptive, context-aware grasping assistance.

\section{Exosuit design and control}
\subsection{Hardware Design}
\begin{figure*}[t]
\centering
\includegraphics[width=0.9\textwidth]{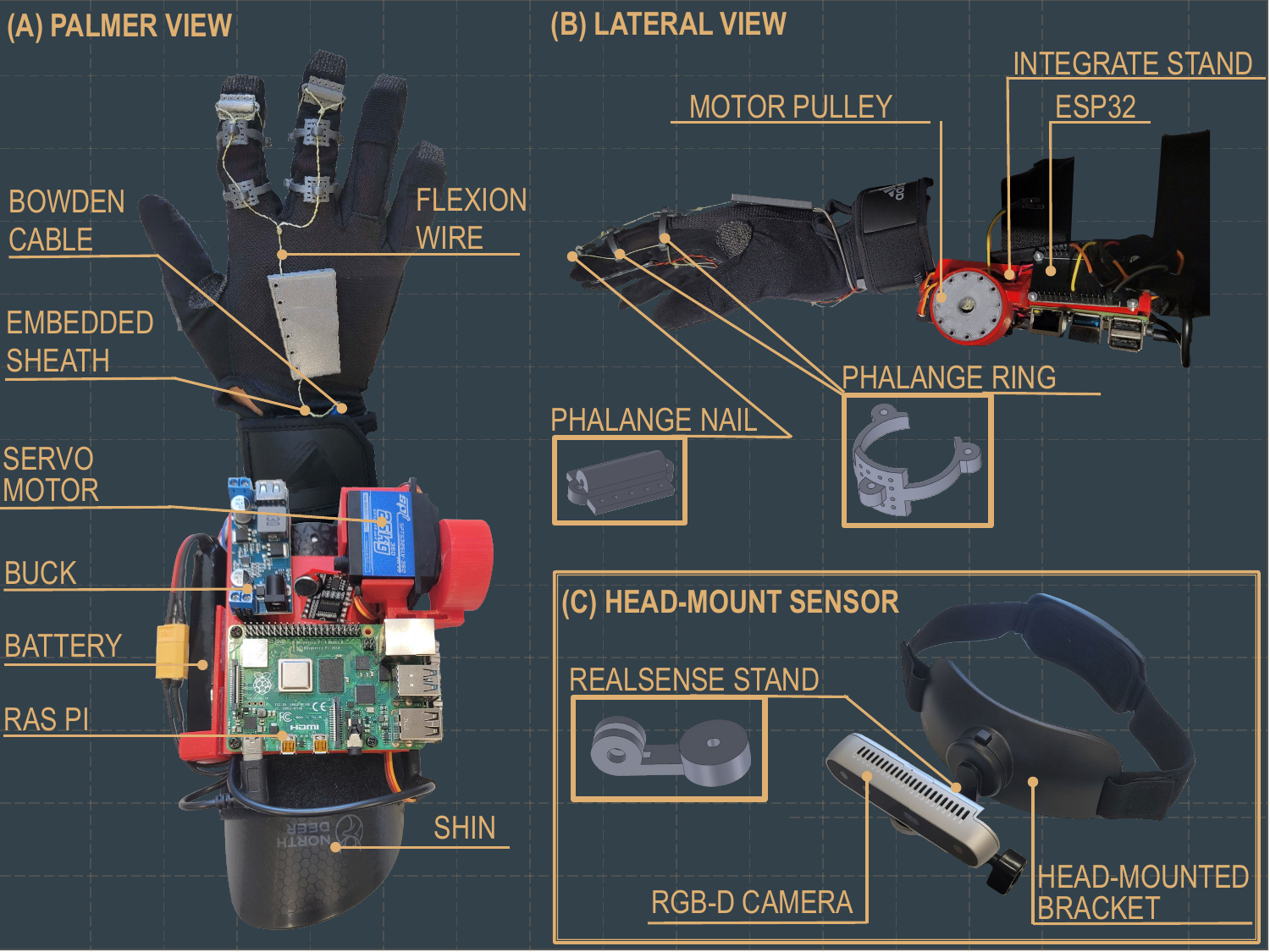}
\caption{Hardware setup of the proposed soft wearable exoskeleton system. (A) \textbf{Palmer view} of the cable-driven glove. (B) \textbf{Lateral view} of the integrated actuator, showing the custom-designed pulley system and ESP32 controller housed on the motor stand. (C) \textbf{Head-mount sensor module} comprising an Intel RealSense RGB-D camera secured to a custom-designed adjustable head-mounted bracket, enabling egocentric visual perception.}
\label{fig:hardware}
\end{figure*}

The implemented soft hand exoskeleton used in the study consisted of three main components (Fig. \ref{fig:hardware}) a cable-driven soft hand exoskeleton based on an existing design \cite{hu2025pointcloudbasedgraspingsoft}; 2) an actuator that integrated the entire control circuit and transferred force from the motor to the finger joints; and 3) an egocentric vision sensing module.

The exoskeleton was composed of flexible polyamide, with 3D-printed rings near the metacarpophalangeal (MCP), proximal interphalangeal (PIP), and distal interphalangeal (DIP) joints, replicating human hand tendons for effective grasping and releasing motions. A cable-driven mechanism connected the actuation system to the glove and was designed to remain fixed, reducing the impact of dynamic sheath bending angles. To maintain a tight, straight connection between the actuation system and the palm-supporting pieces, the sheaths were slightly tensioned, which minimized sheath bending during operation. This approach reduced unwanted contact between the tendon wires and the sheath, effectively stabilizing fingertip force against dynamic changes in the sheath’s bending angle.

The actuation system, powered by a LiPo battery (Crazepony, 1400mAh, 11.1V, 64.1g, Shenzhen, China), weighed 0.5 kg and was mounted on a forearm-worn shin guard (Super Comfortable Shin Pad, Northdeer, China). A flat servo motor (Digital Servo, 4.8V, 24 kgcm, CHICIRIS, China) drove a 30mm diameter pulley, around which the actuation cable was wound. The system employed a microcontroller (Raspberry Pi 4B, 1.5GHz, 4GB RAM, Broadcom, UK) for acquiring depth data, while a secondary chip (ESP32, 2.4GHz, 4GB, Bluetooth, Shanghai, China) managed motor pulse-width modulation (PWM).

To introduce and evaluate the OVGrasp framework, a head-mounted RealSense D415 RGB-D camera (Intel RealSense, California, USA) was incorporated into the soft exoskeleton's original design to capture environmental data. The camera, attached to the hand-mounted stand (Ulanzi, CM027 Head-Mounted Bracket First-Person View Shooting Device, China) via a miniature gimbal, sent depth images through the Raspberry Pi’s Wi-Fi module to a server (Dell, Precision 7680, US) for inference with a 10 frames per second (fps)  frame rate and 640 $\times$ 480 resolution.

To achieve the releasing action after a grasping task, the sensing module incorporated an additional control modality: an automatic speech recognition (ASR) foundation model \cite{gao2023funasr} and a microphone (Ascot City, High Sensitivity Microphone, 3.5 mm, 85 decibels), where actuation commands were transmitted via the corresponding voice control command from users.

\subsection{OVGrasp Control Strategy}
The proposed OVGrasp control strategy provides a hierarchical framework that integrates visual perception, multimodal intent detection, and precise motor control to facilitate effective grasping assistance. It comprises three distinct but interconnected modules: the High-Level Detector, the Middle-Level Decision-Maker, and the Low-Level PID Controller. 

\subsubsection{High-Level Detector}
The high-level detector (Fig. \ref{fig:overview}, purple) employs the YOLO-World architecture \cite{cheng2024yoloworld}, leveraging joint vision-language representation learning for real-time, open-vocabulary object detection. This detector processes RGB frames through a YOLO backbone, generating multi-scale visual features subsequently integrated with textual embeddings obtained from a pre-trained Weight YOLO-World-X \cite{cheng2024yoloworld} via a Vision-Language Path Aggregation Network (PAN). A Box Head outputs bounding boxes, while a Text Contrastive Head computes object embeddings for precise region-text matching. The similarity $s_{k,j}$ between visual embeddings $e_k$ and textual embeddings $w_j$ is computed as:

\begin{equation}
    s_{k,j} = \alpha \cdot L_2(e_k) \cdot L_2(w_j)^\top + \beta,
\end{equation}

where $L_2(\cdot)$ denotes $L_2$-normalization, and $\alpha$, $\beta$ are learnable parameters. The module outputs bounding-box center coordinates to the middle-level decision-maker.

\subsubsection{Middle-Level Decision-Maker}
The middle-level decision-maker (Fig. \ref{fig:overview}, yellow) integrates multimodal visual, depth, and speech inputs to determine grasping intent and translate it into symbolic control commands. The inputs include bounding-box center coordinates $(u, v)$ from the High-Level Detector, a depth frame from the depth sensor, and speech signals. Depth frames are first spatially registered with RGB images, ensuring accurate correspondence. A grasp-point calculator extracts the depth $d$ at each detected object's center, forming nodes represented as $(u, v, d)$.

These nodes form a dynamic Grasp-Point Graph $\mathcal{G}=(\mathcal{V},\mathcal{E})$, where vertices $v_i\in\mathcal{V}$ represent grasp points and edges $\mathcal{E}$ represent spatial proximity between objects. The Euclidean distance $\Delta_i$ between each graph node $v_i=(u_i,v_i,d_i)$ and the hand exoskeleton centroid $h=(u_h,v_h,d_h)$ is calculated as:
\begin{equation}
    \Delta_i = \sqrt{(u_i - u_h)^2+(v_i - v_h)^2+(d_i - d_h)^2}.
\end{equation}

The node with minimal distance $\Delta_{\text{min}}$ is identified as the current target object. To ensure decision stability, a queue tracks the target node across consecutive frames. If the same node appears consistently for $\tau$ frames, a symbolic command 'G' is triggered and transmitted to the low-level controller via Bluetooth. The Symbol System module handles this decision, ensuring commands are only issued when confidence is sufficient.

\begin{algorithm}[t]
\caption{Middle-Level Multimodal Intent Detection}
\label{alg:middle_level}
\SetKwInOut{Input}{Input}
\SetKwInOut{Output}{Output}

\Input{
    RGB frame $I_{rgb}$, Depth frame $I_{depth}$, Speech stream $S$;\\
    Exoskeleton centroid $h = (u_h, v_h, d_h)$; \\
    Queue length threshold $\tau$
}
\Output{
    Symbolic command $C \in \{\texttt{G}, \texttt{R}, \texttt{S}, \emptyset\}$
}

\BlankLine
\tcp{Visual-Depth Branch}
$B \gets$ HighLevelDetector($I_{rgb}$)\;
$I_{depth}^{reg} \gets$ DepthRegister($I_{depth}, I_{rgb}$)\;
\ForEach{object $i$ in $B$}{
    $(u_i, v_i) \gets$ center of bbox $i$\;
    $d_i \gets I_{depth}^{reg}(u_i, v_i)$\;
    $\Delta_i \gets \sqrt{(u_i - u_h)^2 + (v_i - v_h)^2 + (d_i - d_h)^2}$\;
}
$v^* \gets \arg\min_i \Delta_i$ \tcp*{Nearest object}
UpdateQueue($v^*$)\;
\If{SameTargetInQueue($\tau$)}{
    $C \gets \texttt{G}$ \tcp*{Send grip command}
}

\BlankLine
\tcp{Speech Branch}
$T \gets$ FunASR($S$)\;
\If{\texttt{``release''} $\in T$}{
    $C \gets \texttt{R}$\;
}
\ElseIf{\texttt{``stop''} $\in T$}{
    $C \gets \texttt{S}$\;
}

\BlankLine
\Return $C$
\end{algorithm}

Simultaneously, speech signals from the microphone undergo analog-to-digital conversion and are processed by the FunASR model \cite{gao2023funasr}, an advanced end-to-end automatic speech recognition framework leveraging transformer-based encoder-decoder architectures and self-attention mechanisms for robust transcription. FunASR offers improved accuracy, robustness to noise, and efficient inference compared with conventional ASR systems. Extracted textual commands undergo semantic analysis, and detection of keywords initiates specific control commands: identification of the term \textit{“release”} triggers the transmission of 'R', while the detection of \textit{“stop”} initiates the 'S' command, facilitating interactive and intuitive user control.

\subsubsection{Low-Level Controller}
The low-level controller (Fig. \ref{fig:overview}, blue) translates symbolic commands (`G' for grip, `R' for release, and `S' for stop) into accurate actuation commands to drive the cable-driven soft hand exoskeleton. Upon receiving a symbolic command, the controller specifies a user-defined target velocity setpoint ($\Delta v$). A proportional-integral-derivative (PID) feedback control loop modulates the motor’s velocity by dynamically adjusting the pulse-width modulation (PWM) duty-cycle control signal. Formally, the PID control law for velocity regulation is defined as:

\begin{equation}
u(t) = K_p e_v(t) + K_i \int_0^t e_v(\tau)\,d\tau + K_d \frac{d e_v(t)}{dt},
\end{equation}

\noindent where $u(t)$ is the PWM output signal, $e_v(t)$ represents the instantaneous error between the desired velocity setpoint and the actual measured motor velocity, and $K_p$, $K_i$, and $K_d$ denote proportional, integral, and derivative gain parameters, respectively.

Torque regulation is indirectly achieved by imposing a voltage constraint on the motor, which inherently limits the current flow and thereby bounds the maximum attainable torque. This indirect torque control strategy ensures both user safety and mechanical compliance during grasping. Collectively, this dual-layered control approach—explicit PID-based velocity tracking combined with implicit torque limitation—enables smooth, stable, and compliant operation of the exoskeleton, enhancing grasp precision and reliability in practical scenarios.

\section{Study protocol and evaluation}
\subsection{Dataset}
\begin{figure*}[h]
\centering
\includegraphics[width=0.8\textwidth]{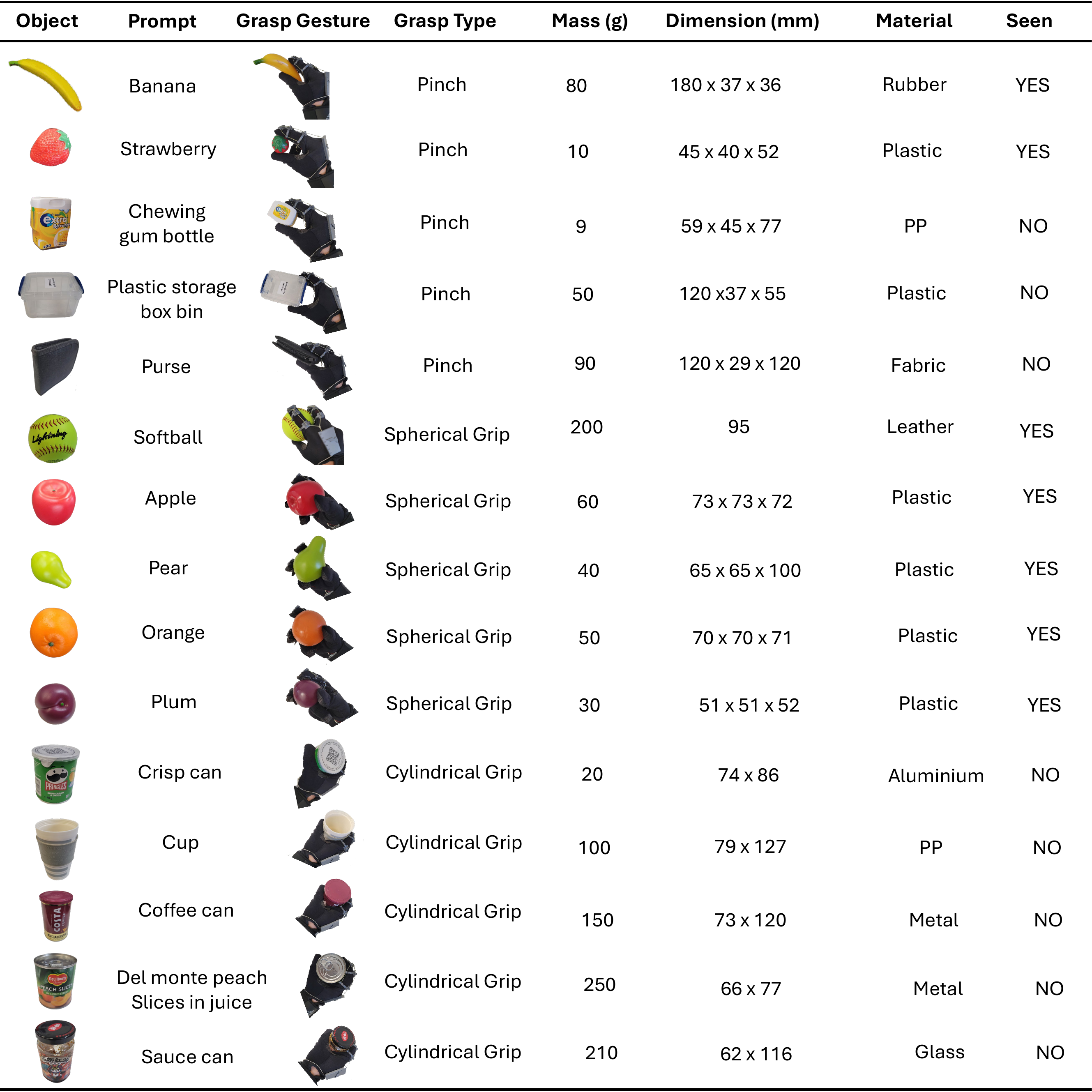}
\caption{Overview of the 15 objects used in the grasping experiments, along with their corresponding prompts, grasp gestures, grasp types, physical properties, and materials. The selected objects span three common grasp types—pinch, spherical, and cylindrical—based on their shape, size, and functional use.}
\label{fig:dataset}
\end{figure*}

The dataset consists of 15 objects in Figure \ref{fig:dataset}, comprising 7 items (banana, strawberry, softball, apple, pear, orange, and plum) selected from the Yale-CMU-Berkeley (YCB) object dataset \cite{calli2015benchmarking}, designated as \textit{'seen'} objects. These \textit{seen} categories are included in the image-text pairs used to pre-train the YOLO-World model \cite{cheng2024yoloworld}. In contrast, 8 additional common household objects (chewing gum box, small storage box, purse, small chips can, cup, coffee can, peach can, and chilli can), designated as \textit{'unseen'}, are not present in the training set of YOLO-World and were randomly selected from daily life to evaluate the model's zero-shot generalization capability. All items were selected according to three distinct grasp categories—pinch, spherical grip, and cylindrical grip—as described in Calli et al.~\cite{calli2015benchmarking}, ensuring diversity in shape, mass, size, and material properties across the dataset.

\subsection{Participants}
To evaluate the grasping performance of the proposed method, ten healthy right-handed participants (eight males and two females; mean age: 25.0 ± 6 years; weight: 75 ± 14 kg; height: 1.80 ± 0.22 m) were recruited. All participants demonstrated normal hand motor function and were instructed to perform the grasping tasks comfortably, ensuring these movements did not cause pain or discomfort.

The study protocol was reviewed and approved by the University and College Research Ethics Committee (CREC) at King's College London (ethical approval number: \textit{MRPP-23/24-40750}). All participants received comprehensive information regarding the study's objectives and procedures and provided written informed consent prior to their participation.

\subsection{Study Processing}
\textbf{Grasping Ability Score:} Once participants agreed to take part, they were seated next to a table, as illustrated in Fig. \ref{fig:overview}. They were instructed to maintain their hands relaxed and avoid exerting any force on the objects during the grasping process. A researcher presented each object to the participants, who were then asked to maintain the grasp for three seconds. Participants subsequently rotated their hands by 180 degrees to achieve a palm-down position, holding the grasp for another three seconds before the system initiated the release. Additionally, the researcher recorded the number of times the camera made contact with the table surface to assess whether the sensor’s positioning underneath the hand caused any discomfort.

In accordance with the methodology proposed by Maldonado et al. \cite{maldonado2023fabric}, each object was grasped three times using OVGrasp to evaluate the vision-based method’s grasping and maintaining capabilities. The evaluation period began when participants attempted to grasp the object and ended when the object was released. The grasping capability (Grasping) was scored as follows: 0 (failed grasp), 0.5 (incorrect grasp), or 1 (correct grasp). Similarly, the maintaining capability (Maintaining) was scored as: 0 (object dropped), 0.5 (object movement), or 1 (object stable). Scores from individual objects within each grasp type were averaged to obtain an overall score per grasp type. A cumulative Grasping Ability Score (GAS) \cite{llop2019anthropomorphic} was then computed to provide a comprehensive evaluation of the proposed vision-based method’s performance.

\textbf{Zero-shot Evaluation:} We evaluate the high-level visual controller YOLO-World on our custom dataset for zero-shot object detection. The dataset includes 15 object categories, among which 8 are designated as unseen. The objects are randomly grouped into five sets, and for each group, a 1-minute video sequence is recorded. From each video, 20 representative keyframes are extracted and used as input for the detector. We report the results in terms of $AP_{seen}$, $AP_{unseen}$, $AP_{all}$, and $mAP$ at $IoU$ $\geq$ 0.5 \cite{everingham2010pascal}. This protocol enables a quantitative assessment of zero-shot generalization performance in realistic assistive scenarios without additional fine-tuning.

\textbf{Kinematics Analysis of Fingers:} The biomechanical impact of the proposed controller on the range of motion (ROM) was assessed through video recordings captured during the experimental tasks. ROM measurements were specifically recorded at key finger joints, metacarpophalangeal (MCP), proximal interphalangeal (PIP), and distal interphalangeal (DIP), while participants performed grasping tasks both with and without the assistance of the hand exoskeleton. Three distinct grasp types were evaluated: cylindrical (C), spherical (S), and pinch (P). To accurately determine ROM values, YOLO11 \cite{yolo11_ultralytics} was fine-tuned using the 11k Hands Dataset \cite{afifi201911kHands} over 100 epochs. The resulting data were analyzed to compare joint ROM under conditions of external actuation versus natural hand movements, thereby providing a comprehensive evaluation of the controller's biomechanical effectiveness.

\section{Experimental results}
\subsection{Grasping Ability Score}
Table \ref{tab:gas} summarizes the average grasping performance across three grasp types (pinch, spherical, and cylindrical) based on the Grasping Ability Score (GAS) \cite{llop2019anthropomorphic}, which is decomposed into grasping and maintaining components. The proposed vision-based approach achieved state-of-the-art GAS scores for both spherical and cylindrical grips, outperforming existing push-button and force-sensing baselines.

\begin{table}[h]
\centering
\renewcommand{\arraystretch}{1.5}
\caption{Grasping performance averages divided into Grasping, Maintaining, and total GAS scores for the three types of grasp.}
\label{tab:gas}
\resizebox{\columnwidth}{!}{%
\begin{tabular}{l l c c c}
\hline
\textbf{Method}        & \textbf{GAS (\%)}      & \textbf{Pinch ($\uparrow$)}            & \textbf{Spherical Grip ($\uparrow$)}   & \textbf{Cylindrical Grip ($\uparrow$)} \\ \hline
\textbf{Push-button \cite{hu2025pointcloudbasedgraspingsoft}}    & Grasping score          & 94.67 $\pm$ 0.72          & 88.00 $\pm$ 1.89         & 84.00 $\pm$ 1.84         \\
~                      & Maintaining score       & 89.33 $\pm$ 1.19          & 74.33 $\pm$ 3.12         & 92.33 $\pm$ 1.25         \\
~                      & GAS score               & 92.00 $\pm$ 0.95          & 81.16 $\pm$ 2.1          & 74.16 $\pm$ 2.29         \\ \hline
\textbf{Force-sensing\cite{hu2025pointcloudbasedgraspingsoft}} & Grasping score          & 96.67 $\pm$ 0.75          & 91.33 $\pm$ 1.89         & 85.33 $\pm$ 1.97         \\
~                      & Maintaining score       & 89.33 $\pm$ 1.59          & 78.00 $\pm$ 2.44         & 91.73 $\pm$ 2.05         \\
~                      & GAS score               & 93.00 $\pm$ 1.17          & 84.66 $\pm$ 2.17         & 78.33 $\pm$ 2.23         \\ \hline
\textbf{Maldonado-Mej\'ia et al.\cite{maldonado2023fabric}} & Grasping score  & 59.44 $\pm$ 0.26          & 75.33 $\pm$ 0.14         & 93.33 $\pm$ 0.78         \\
~                      & Maintaining score       & 93.33 $\pm$ 0.47          & 57.44 $\pm$ 0.74         & 92.22 $\pm$ 0.34         \\
~                      & GAS score               & 76.39 $\pm$ 0.11          & 83.89 $\pm$ 0.26         & 80.28 $\pm$ 0.31         \\ \hline
\textbf{Proposed approach} & Grasping score     & 92.00 $\pm$ 0.34  & \textbf{94.67 $\pm$ 0.32}  & 91.33 $\pm$ 0.43 \\
~                      & Maintaining score      & 83.33 $\pm$ 0.49  & \textbf{84.67 $\pm$ 0.45}  & 76.00 $\pm$ 0.53 \\
~                      & GAS score              & 87.65 $\pm$ 0.42  & \textbf{89.67 $\pm$ 0.77}  & \textbf{83.67 $\pm$ 0.48} \\ \hline
\end{tabular}%
}
\end{table}

Specifically, our method reached a GAS of 89.67 ± 0.77\% for spherical and 83.67 ± 0.48\% for cylindrical grips, significantly improving the comparison methods. Notably, the grasping scores for spherical grips (94.67 ± 0.32\%) were also the highest among all methods tested.

Qualitative user feedback suggests that the proposed vision-based triggering mechanism allowed participants to focus entirely on aligning the hand with the optimal grasping region. This is attributed to the system's context-aware control, which more naturally aligns with human grasping behavior, automatically initiating closure when approaching the object. In contrast, the push-button method \cite{hu2025pointcloudbasedgraspingsoft,maldonado2023fabric} required participants to divide attention between object positioning and using the contralateral hand to press the trigger. Similarly, the force-sensing method \cite{hu2025pointcloudbasedgraspingsoft} demanded physical contact and applied pressure at specific sensor locations, diverting user attention from precise grasp execution.


\begin{table}[t]
\centering
\renewcommand{\arraystretch}{1.5}
\caption{Comparison of Grasping Ability Scores (GAS) for individual objects under eye-in-hand and egocentric visual perspectives. Results are grouped by grasp type (pinch, spherical, cylindrical).}
\label{tab:gas_object}
\resizebox{\columnwidth}{!}{%
\begin{tabular}{l l c c}
\hline
\textbf{Grip Type}        & \textbf{Object}      & \textbf{Eye-in-hand View GAS (\%) \cite{hu2025pointcloudbasedgraspingsoft}}            & \textbf{Egocentric view GAS (\%)}  \\ \hline
\textbf{Pinch} & Banana           & 94.16 $\pm$ 1.51  & \textbf{95.00 $\pm$ 1.25}  \\
~              & Strawberry       & 96.67 $\pm$ 0.86  & 92.50 $\pm$ 1.94   \\
~              & Chewing gum      & 100.00 $\pm$ 0.00 & 100.00 $\pm$ 0.00  \\
~              & Storage box      & 98.33 $\pm$ 0.43  & 88.34 $\pm$ 2.15  \\
~              & Purse            & 100.00 $\pm$ 0.00 & 61.66 $\pm$ 0.43  \\ \hline
\textbf{Spherical} & Softball     & 100.00 $\pm$ 0.00 & 100.00 $\pm$ 0.00  \\
~              & Apple            & 83.34 $\pm$ 1.72  & 82.50 $\pm$ 2.8  \\
~              & Pear             & 97.50 $\pm$ 0.65  & 96.67 $\pm$ 0.86  \\
~              & Orange           & 79.17 $\pm$ 1.94 & 72.50 $\pm$ 1.94  \\
~              & Plum             & 97.50 $\pm$ 0.65  & 96.67 $\pm$ 0.86   \\ \hline
\textbf{Cylindrical} & Chips can  & 65.00 $\pm$ 3.01  & \textbf{69.17 $\pm$ 3.66}  \\
~              & Coffee can       & 75.84 $\pm$ 3.66  & 75.00 $\pm$ 3.01  \\
~              & Cup              & 100.00 $\pm$ 0.00  & 100.00 $\pm$ 0.00  \\
~              & Peach can        & 80.00 $\pm$ 2.58  & \textbf{83.34 $\pm$ 1.72}  \\
~              & Sauce can        & 99.16 $\pm$ 0.22   & 90.83 $\pm$ 1.51  \\ \hline
\end{tabular}%
}
\end{table}

Table \ref{tab:gas_object} presents a comparative analysis of the GAS achieved under egocentric and eye-in-hand visual perspectives across 15 objects categorized by grip types. The experiments were conducted with the same tendon-driven soft hand exoskeleton worn by 10 participants, ensuring consistent mechanical structure and control conditions.

For pinch grasps, the eye-in-hand view yielded slightly higher overall performance across most objects. Comparable GAS scores were observed for banana, strawberry, chewing gum, and storage box. However, the egocentric view showed a notable performance drop in grasping the purse (61.66 ± 0.43\%), likely due to visual ambiguity between the purse and the user’s clothing or background, as both shared similar fabric texture and color, occasionally failing to trigger grasp detection. Conversely, the egocentric view outperformed eye-in-hand for the banana, which has a relatively low profile. The eye-in-hand camera, mounted beneath the hand, restricted the user’s ability to approach optimal grasping angles due to a collision risk with the tabletop.

In the spherical grip category, both viewpoints achieved similar GAS scores across all objects. This consistency can be attributed to the grasping geometry—objects were tall enough to provide clearance between the hand and the table, minimizing interference from the downward-facing camera. As a result, the shared mechanical and control structure produced comparable grasping performance across both views.

For cylindrical grasps, the egocentric view demonstrated superior GAS, particularly for low-profile objects such as a crisp can and a peach can. In the eye-in-hand configuration, the under-wrist camera limited hand positioning in close proximity to short objects, affecting grasp alignment. By contrast, the egocentric camera provided a clear top-down perspective without introducing physical obstructions, thereby improving target alignment and grasp reliability.

\subsection{Kinematics Analysis of Fingers}

\begin{figure*}[h]
\centering
\includegraphics[width=1\textwidth]{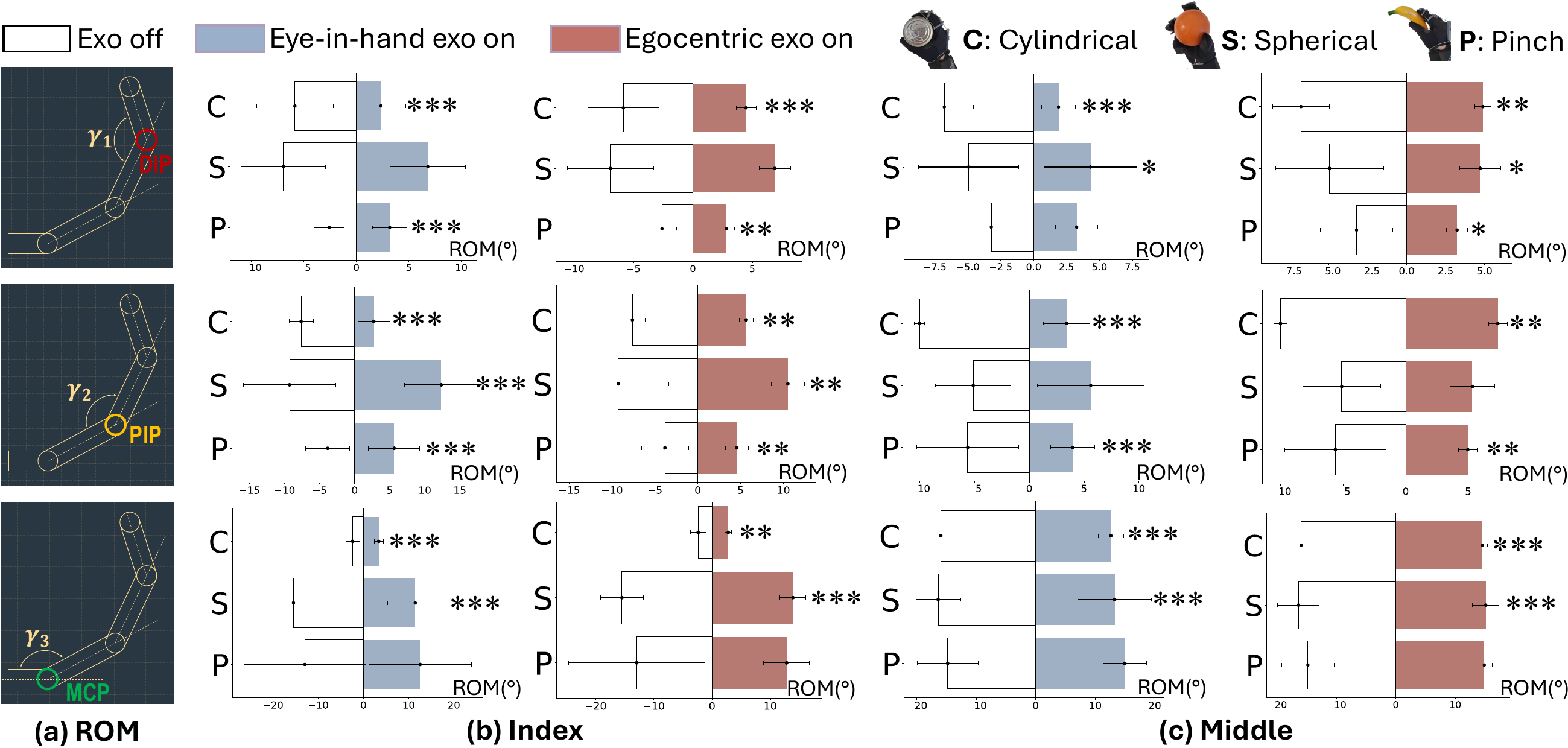}
\caption{Range of motion (ROM) analysis for DIP, PIP, and MCP joints of the index and middle fingers during grasping tasks using cylindrical (C), spherical (S), and pinch (P) grips. Ten participants performed grasps on 15 objects under three conditions: without the exoskeleton (white bars), with the exoskeleton using an eye-in-hand camera (blue bars), and with the exoskeleton using an egocentric camera (red bars). Significance levels are indicated by asterisks (*), where * denotes p $\leq$ 0.05, ** denotes p $\leq$ 0.01, and *** denotes p $\leq$ 0.001.}
\label{fig:hand_analysis}
\end{figure*}

Figure \ref{fig:hand_analysis} illustrates the range of motion (ROM) \cite{bland2008restricted} at the distal interphalangeal (DIP), proximal interphalangeal (PIP), and metacarpophalangeal (MCP) joints of the index and middle fingers during grasping tasks with and without the hand exoskeleton, across three grasp types (cylindrical, spherical, and pinch). Ten participants performed grasps on 15 different objects using both eye-in-hand (blue) and egocentric (red) visual configurations. Bar plots represent ROM differences between the exoskeleton-on and exoskeleton-off conditions, and statistical significance was assessed using p-values.

For the DIP joint, the egocentric configuration showed significantly more natural ROM in cylindrical and pinch grasps, more closely replicating unassisted hand movement. Both configurations demonstrated consistent improvements in spherical grasps. At the PIP joint, the egocentric view consistently yielded greater ROM alignment with natural grasping in both index and middle fingers across all grasp types. This indicates that the egocentric configuration more effectively preserves intrinsic joint kinematics during assisted motion. The MCP joint exhibited generally similar behavior across both views, with the exception of the index finger during cylindrical grasps, where eye-in-hand control showed slightly reduced ROM significance.

Overall, the egocentric viewpoint appeared to reduce the kinematic interference introduced by the exoskeleton, thereby enabling movement patterns that more closely approximate natural, unassisted grasping, particularly in the distal joints and across more constrained grasp types.

\subsection{Ablation Study}
To evaluate the impact of the open-vocabulary detection (OVD) mechanism in our high-level visual pipeline, we conducted a comparative study between a standard YOLOv8 model \cite{yolov8_ultralytics} and its OVD-integrated counterpart (YOLO-World \cite{cheng2024yoloworld}). Both models were pretrained on identical image-text datasets (O365 \cite{shao2019objects365}, GoldG \cite{zhai2022lit}, and CC3M \cite{sharma2018conceptual}) to ensure a fair comparison.

\begin{table}[h]
    \centering
    \renewcommand{\arraystretch}{1.5}
    \caption{Zero-shot Evaluation on Seen and Unseen Objects. We report $AP$ on seen, unseen, and all categories, along with $mAP$, to compare YOLO models with and without open-vocabulary detection (OVD). Both models are pretrained on O365+GoldG+CC3M.}
    \label{tab:ablation_ov}
    \resizebox{\columnwidth}{!}{
    \begin{tabular}{lccccc}
        \hline
        \textbf{Methods} & Pre-trained Data &      $AP_{seen}$ &  $AP_{unseen}$ &  $AP_{all}$ &  $mAP$ \\ \hline
        YOLO without OVD   & O365+GoldG+CC3M &   1.0 & 0.86 & 0.90 & 0.93  \\ \hline
        YOLO with OVD      & O365+GoldG+CC3M &   0.43 & 0.03 & 0.13 & 0.23 \\ \hline
    \end{tabular}
    }
\end{table}

As shown in Table \ref{tab:ablation_ov}, the OVD-enabled detector achieves substantially higher performance on unseen object categories. Specifically, YOLO with OVD achieves an $AP$ of 93.3\% compared to 33.3\% by YOLO without OVD. The $AP$ on unseen categories ($AP_{unseen}$) rises sharply from 0.03 to 0.86, demonstrating enhanced zero-shot generalization. Additionally, the model achieves higher $AP_{all}$ (0.90 vs. 0.13) and $mAP$ (0.93 vs. 0.23), underscoring the effectiveness of prompt-based detection in open environments. These findings validate that integrating a language-guided open-vocabulary mechanism significantly improves the contextual perception and versatility of assistive grasping systems.

\begin{figure*}[h]
\centering
\includegraphics[width=0.8\textwidth]{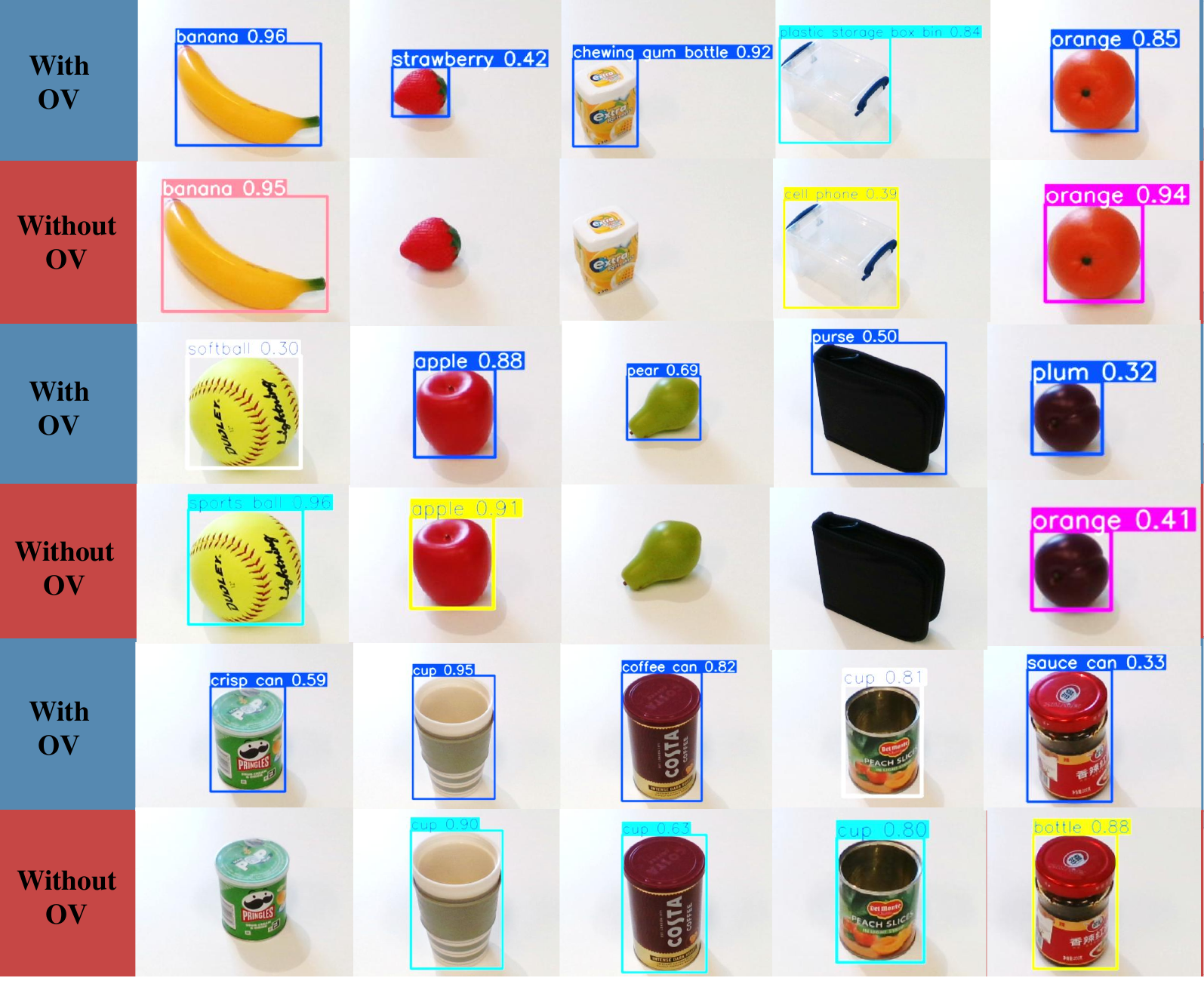}
\caption{Visualization results on zero-shot inference on 7 objects and non-zero-shot Inference on 8 objects with and without the open-vocabulary (OV) mechanism. The top rows show detection results using YOLO with OV (blue), while the bottom rows show results from the baseline YOLO without OV (red). The OV-enabled detector successfully identifies 8 previously unseen objects using prompt-based embeddings, demonstrating superior generalization across varied object appearances, including small (e.g., strawberry), low-profile (e.g., crisp can), deformable (e.g., purse), and uncommon items (e.g., chewing gum bottle).
}
\label{fig:ablation_ov}
\end{figure*}

To further illustrate the qualitative differences, Figure \ref{fig:ablation_ov} visualizes detection outputs from the two detectors under identical scenes. Blue-bordered boxes (With OVD) denote detections from the YOLO-World model, while red-bordered boxes (Without OVD) indicate results from the baseline YOLOv8. The OVD-enhanced detector demonstrates superior performance, particularly in identifying small (e.g., strawberry), short (e.g., crisp can), irregularly shaped (e.g., pear), and uncommon objects (e.g., chewing gum bottle, fabric purse) that are typically underrepresented in conventional detection datasets. This aligns with our design goal of enhancing situational awareness for wearable robotic systems operating in open environments.

In summary, these findings highlight the practical significance of open-vocabulary grounding in wearable exoskeleton applications. The ability to flexibly recognize diverse objects, guided only by user-defined language prompts, enables context-adaptive grasp assistance, especially for stroke survivors and users with impaired hand function who engage with varied household objects beyond the scope of conventional datasets.

\section{Conclusion and future works}
In this paper, we presented \textbf{OVGrasp}, a hierarchical and multimodal control framework designed to enhance grasping assistance for wearable soft exoskeletons in open-world scenarios. By incorporating an open-vocabulary vision-language model, OVGrasp enables zero-shot generalization to novel objects without requiring additional fine-tuning, addressing a key limitation in traditional perception-driven assistive systems. We further introduced a multimodal intent detection module that integrates visual and speech signals to support multi-object selection and dynamic grasp–release interaction. OVGrasp was implemented on a custom-built, cable-driven soft hand exoskeleton equipped with an egocentric RGB-D camera. Extensive studies demonstrated that OVGrasp achieves a state-of-the-art Grasping Ability Score (GAS) of 87.00 ± 2.46\% compared to push-button and force-sensing controllers and improves finger joint kinematic profiles compared to eye-in-hand baselines.

Future research will further explore the use of egocentric vision in wearable grasp-assistive systems. This viewpoint more closely aligns with the natural way humans perceive and act upon their environment, while eliminating physical interference from hand-mounted cameras, thereby enabling more intuitive and unobstructed grasping behavior in daily activities. In addition, future work will address limitations in handling transparent and deformable objects, which remain challenging due to depth sparsity and context ambiguity. We aim to improve scene understanding by distinguishing between rigid and deformable targets, and to adapt motor control through precision modulation of the torque PID loop.

Another promising direction involves the integration of high dynamic range visual sensors, such as event-based cameras, into wearable robotic gloves to enhance responsiveness and robustness under challenging lighting conditions. Finally, we plan to validate the full system through clinical trials with individuals affected by upper-limb impairments, to promote independent living and improve quality of life.

\bibliographystyle{elsarticle-num} 
\bibliography{reference}

\end{document}